\documentclass{article}
\usepackage{spconf,graphicx, cite}
\usepackage{float}
\usepackage{ragged2e}
\usepackage{enumitem}
\usepackage[caption=false]{subfig}
\usepackage{pifont}
\usepackage{amsmath,amssymb} 
\usepackage[linesnumbered,ruled,vlined]{algorithm2e}
\usepackage[table, svgnames, dvipsnames]{xcolor}
\usepackage{bm}      
\usepackage{comment}

\title{Sparse Mixture Once-for-all Adversarial Training for Efficient In-Situ Trade-Off Between Accuracy and Robustness of DNNs}
\name{\begin{tabular}{c}Souvik Kundu$^{1}$ $^{2}$, Sairam Sundaresan$^{1}$, Sharath Nittur Sridhar$^{1}$, \\Shunlin Lu$^{2}$, Han Tang$^{2}$, Peter A. Beerel$^{2}$\end{tabular}}
\address{$^{1}$Intel Labs, San Diego, USA\\$^{2}$University of Southern California, Los Angeles, USA}
\begin{document}
%
\maketitle
\begin{abstract}
Existing deep neural networks (DNNs) that achieve state-of-the-art (SOTA) performance on both clean and adversarially-perturbed images rely on either activation or weight conditioned convolution operations. However, such conditional learning costs additional multiply-accumulate (MAC) or addition operations, increasing inference memory and compute costs. 
To that end, we present a \textit{\underline{s}parse \underline{m}ixture once for all \underline{a}dversa\underline{r}ial \underline{t}raining} (SMART), that allows a model to train once and then in-situ trade-off between accuracy and robustness, that too at a reduced compute and parameter overhead. In particular, SMART develops two expert paths, for clean and adversarial images, respectively, that are then conditionally trained via respective dedicated sets of binary sparsity masks. Extensive evaluations on multiple image classification datasets across different models show SMART to have up to $2.72\times$ fewer non-zero parameters costing proportional reduction in compute overhead, while yielding SOTA accuracy-robustness trade-off. Additionally, we present insightful observations in designing sparse masks to successfully condition on both clean and  perturbed images.
\end{abstract}
\section{Introduction}
\label{sec:intro}
The growing use of deep neural networks (DNNs) in safety-critical applications \cite{bojarski2016end,han2021advancing} has intensified the demand for high classification accuracy on both clean and adversarially-perturbed images \cite{wang2020once}. Of the various defense mechanisms \cite{kundu2022towards, he2019parametric} against gradient-based image perturbations \cite{madry2017towards}, adversarial training \cite{madry2017towards,goodfellow2014explaining} has proven to be a consistently effective method to achieve SOTA robustness.  
However, adversarial training causes a significant drop in clean-image accuracy \cite{tsipras2018robustness}, highlighting an accuracy-robustness trade-off that has been explored both theoretically and experimentally \cite{sun2019towards}. This limitation creates a significant roadblock in resource-constrained applications where the users' choice of focus on robustness or clean accuracy depends on  the context \cite{wang2020once, kundu2022fast}. 

Recently, some research \cite{wang2020once, kundu2022fast} has proposed conditional learning of models that provides an in-situ trade-off between accuracy and robustness without incurring the training or storage cost of multiple models. In particular, these works use activation (OAT) \cite{wang2020once} or weight (FLOAT) \cite{kundu2022fast} conditioning to in-situ transform and propagate the activation outputs suitable for the desired user selected trade-offs. However, to yield SOTA accuracy on both clean and perturbed images, the activation-conditioned models require significant increases in storage, training, and inference latency \cite{kundu2022fast}. Weight conditioning mitigates most of these issues, however, significantly increases the costs associated with adding noise to the weights. Interestingly, with the reduced bit-width, that are often sufficient for DNN inference, the difference in energy consumption of MAC and ADDs is significantly reduced. For example, the gap in MAC and ADD energy reduces by $4.2\times$ as we move from 32-b to 8-b INT operation \cite{horowitz20141,kundu2022bmpq}. More importantly, recent advancements of shift-add networks \cite{you2020shiftaddnet} replaces MACs with shift-add that are only $1.8\times$ costlier than ADD, potentially making naive deployment of FLOAT significantly increase the compute budget compared to baseline inference models.
\begin{figure}[!t]
\includegraphics[width=0.40\textwidth]{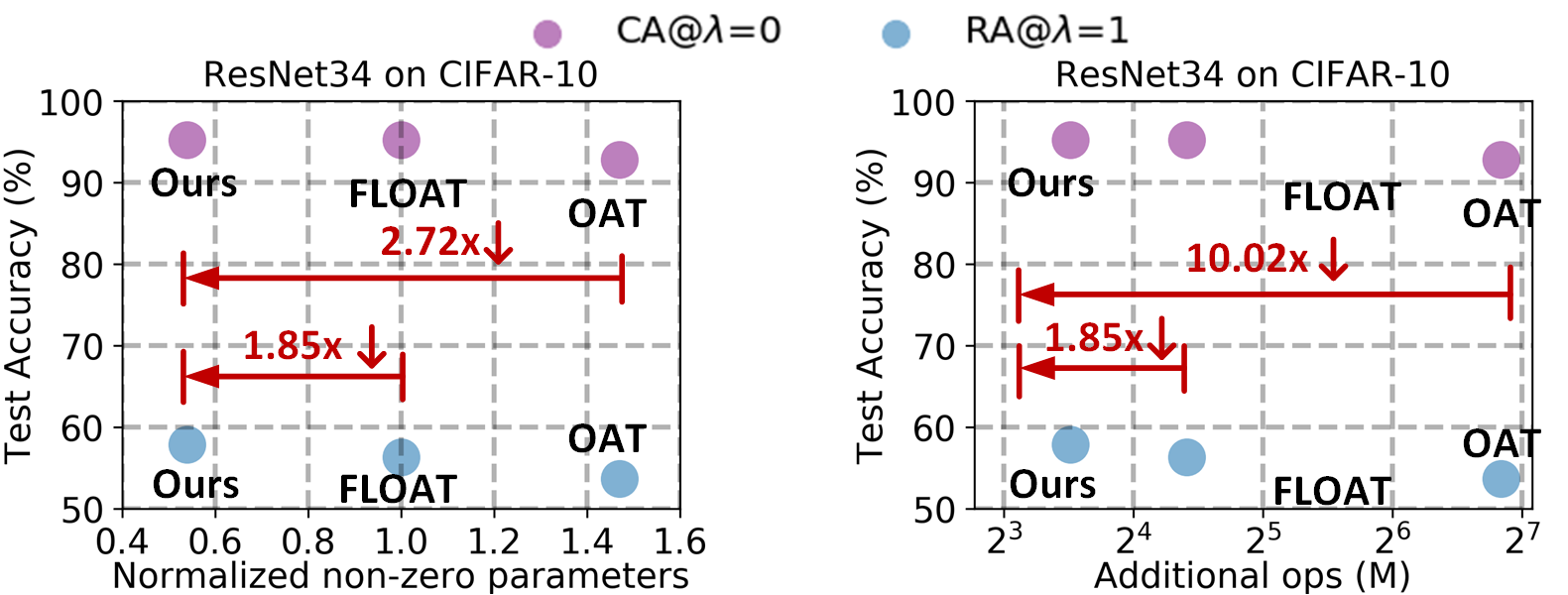}
\vspace{-4mm}
\centering
   \caption{Performance comparison of the proposed method with the existing SOTAs. Here, we report the best clean accuracy (CA) and robust accuracy (RA) of the models trained via respective methods. For the parameter count, we report normalized values with that of a ResNet34 model. Note, here we have converted 1 FLOP to the equivalent ADD operation (op) in 8-b INT format.}
\label{fig:intro_summary}
\vspace{-6mm}
\end{figure}

\textbf{Our contributions.} Our contributions are two-fold. For a residual model (ResNet18), we first analyze the layer-wise importance of model parameters. Based on our analysis, for a model, at later layers, we allow two different sets of sparse masks for clean and adversarial images to create \textit{dedicated expert paths}, while enabling shared dense weights at the initial layers.  We class this \textit{\underline{s}parse \underline{m}ixture once for all \underline{a}dversa\underline{r}ial \underline{t}raining} (SMART) because of the mixing of two sparse expert paths, allowing a single model to train once yet providing different priority trade-offs between CA and RA during inference. We further provide empirical observations as guideline to better design the sparse masks. Extensive experiments with ResNet and wide ResNet model variants on CIFAR-10, CIFAR-100, STL10, and Tiny-ImageNet show SMART models can provide a reduced compute overhead by up to $1.85\times$ while having similar or better CA-RA trade-off (Fig. \ref{fig:intro_summary}).       
\section{Preliminaries}
\label{sec:prelim}
\begin{figure}[!t]
\includegraphics[width=0.33\textwidth]{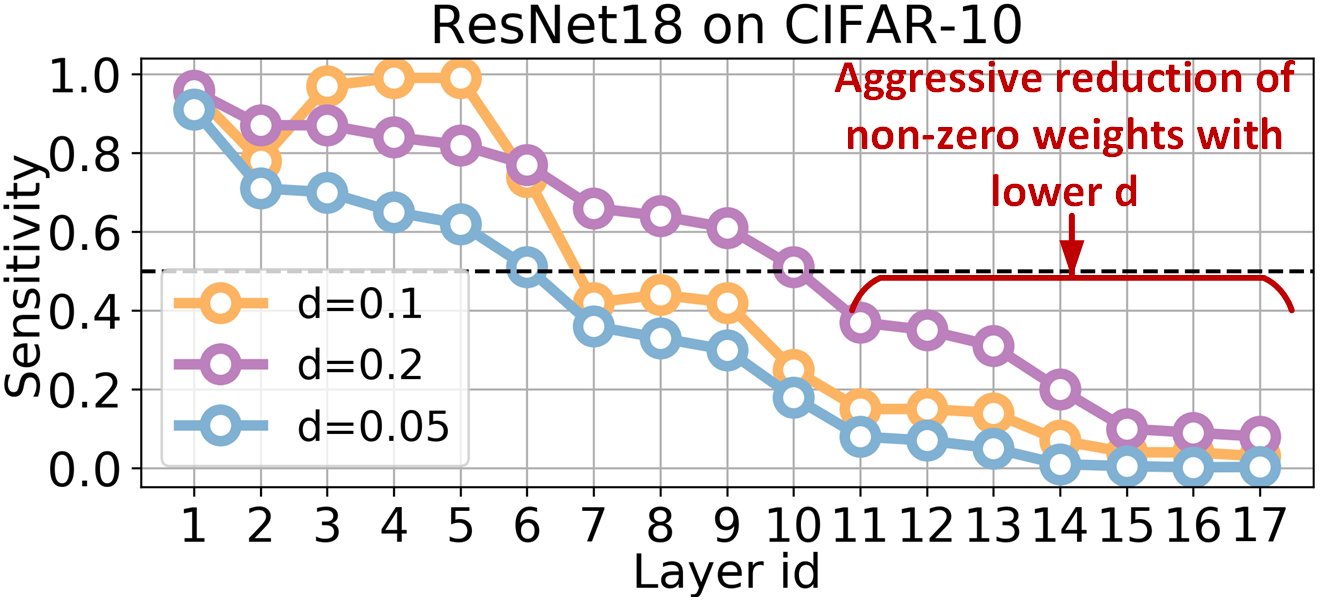}
\vspace{-4mm}
\centering
   \caption{Layer-wise sensitivity analysis of ResNet18 for different global parameter densities ($d$). As we go deeper into the model, the layer's sensitivity reduces letting us set more parameters to zero while maintaining  accuracy.}
\label{fig:motive_sparse}
\vspace{-4mm}
\end{figure}
\subsection{Adversarial Training}
\label{subsec:at}
Adversarial training (AT) aims to train a model with both clean and adversarially perturbed variants of the training images \cite{madry2017towards} in improving model robustness against gradient-based attacks \cite{goodfellow2014explaining}. In particular, projected gradient descent (PGD) attack \cite{madry2017towards}, a strong $L_{\infty}$ perturbation generation algorithm, is widely adopted to create adversarial images during training. The perturbed image for a PGD-$k$ attack with $k$ as the number of steps is given by
\vspace{-2mm}
\begin{align}
\hat{\bm {x}}^{k}&=\texttt{Proj}_{P_{\epsilon}(\bm{x})} (\hat{{\bm x}}^{k-1} + \sigma \times \texttt{sign}(\nabla_{x}\mathcal{L}(f_{\Phi}(\hat{\bm x}^{k-1}, \bm{\Theta}; t))) 
\label{eq:pgd}
\end{align}
The severity of perturbation is controlled by the scalar $\epsilon$. The projection function $\texttt{Proj}$ projects the updated sample onto the projection space $P_{\epsilon}(\bm{x})$. The space itself is the $\epsilon$-$L_{\infty}$ neighbourhood of the benign sample  $\bm{x}$\footnote{Note that the generated $\hat{\bm {x}}$ are clipped to a valid range which, for our experiments, is $[0,1]$.}. Finally, $\sigma$ is the attack step-size. For PGD-AT, the model parameters are then learned by the following ERM 
\vspace{-2mm}
\begin{align}
     [\underbrace{(1-\lambda)\mathcal{L}(f_{\Phi}(\bm{x}, \bm{\Theta}; t))}_{\mathcal{L}_C} + \underbrace{\lambda\mathcal{L}(f_{\Phi}(\hat{\bm{x}}, \bm{\Theta}; t))}_{\mathcal{L}_A}],
\label{eq:adv_loss}
\vspace{-4mm}
\end{align}
Here, $\mathcal{L}_C$ and $\mathcal{L}_A$ are the clean and adversarial loss components, and $\lambda$ is a weighting factor. Hence, for a fixed $\lambda$ and adversarial strength, the model learns a fixed trade-off between accuracy and robustness. For example, an AT with $\lambda$ value of $1$ will allow the model to completely focus on perturbed images, resulting in a significant drop in clean-image classification accuracy. Model robustness can also be improved through the addition of noise to the model's weight tensors. For example, \cite{he2019parametric} introduced the idea of noisy weight tensors with a learnable noise scaling factor. While this showed improved robustness against gradient-based attacks, it incurs a significant drop in clean accuracy.

\subsection{Conditional Learning}
\label{subsec:cond_learn}
Conditional learning trains a model through multiple computational paths that are selectively 
enabled during inference \cite{wang2018skipnet}. These can be early exit branches at differing depths in the architecture to allow early predictions of various inputs \cite{teerapittayanon2016branchynet,huang2017multi,kaya2019shallow,kundu2021hire} or switchable Batch Normalization layers (BNs)
\cite{yu2018slimmable} that enable the network to adjust the channel widths dynamically. Feature transformation can also be leveraged to modulate intermediate DNN features \cite{yang2019controllable}, \cite{wang2020once}. 
In particular, \cite{wang2020once} used FiLM \cite{perez2018film} to adaptively perform a channel-wise affine transformation after each BN stage controlled by $\lambda$ from Equation \ref{eq:adv_loss}. This can yield models that can provide SOTA CA-RA trade-off over various $\lambda$ choices during inference (dubbed once-for-all adversarial training or OAT). \cite{kundu2022fast}, on the other hand, presented a weight-conditioned learning, to transform the weights by conditionally adding scaled-noise to them. However, these methods incur additional compute, storage, and often latency costs increasing the compute footprint compared to a standard model with no support for conditioning.

\section{Motivational Analysis}
\label{sec:motiv}
Model robustness and accuracy can degrade with parameter reduction \cite{timpl2022understanding,kundu2021dnr,kundu2021analyzing, kundu2020pre}. However, \textit{do all layers need to remain overparameterized during the training or can we keep a few layer parameters frozen to zero throughout the training without hurting accuracy?}
To understand the model \textit{parameter utility} at different layers, we conducted three different experiments to prune a ResNet18 to meet three different parameter density budget ($d$) via sparse learning technique \cite{kundu2021dnr}\footnote{We use sparse learning as it assigns non-zero weights based on layer importance evaluated via various proxies including gradient and Hessian}. Here smaller $d$ means the model will have fewer non-zero parameters. For each layer, we then evaluate the parameter utility as $\frac{\# \text{non-zero parameters of the layer}}{\# \text{total number of parameters in the layer}}$. This term also acts as a proxy to measure layer sensitivity towards final accuracy, meaning the higher the sensitivity the more the utility as it demands the majority of the model parameters to be non-zero.  As shown in Fig. \ref{fig:motive_sparse}, deeper layers can afford to be extremely sparse, while earlier layers require more parameters to be non-zero and tend to push more weights to zero as we reduce $d$. Also, as the figure shows, the $\%$ parameter utility of the later layers remains $<50\%$. This is particularly important, as it clearly indicates that a mode can safely keep a large fraction of its weights at zero, yet retain classification performance. Note here, we chose maximum $d=0.2$ as generally density of $0.2$ is sufficient to retain similar accuracy as the baseline. 

\section{SMART Methodology}
\label{sec:proposed_framework}
\begin{figure}[!t]
\includegraphics[width=0.49\textwidth]{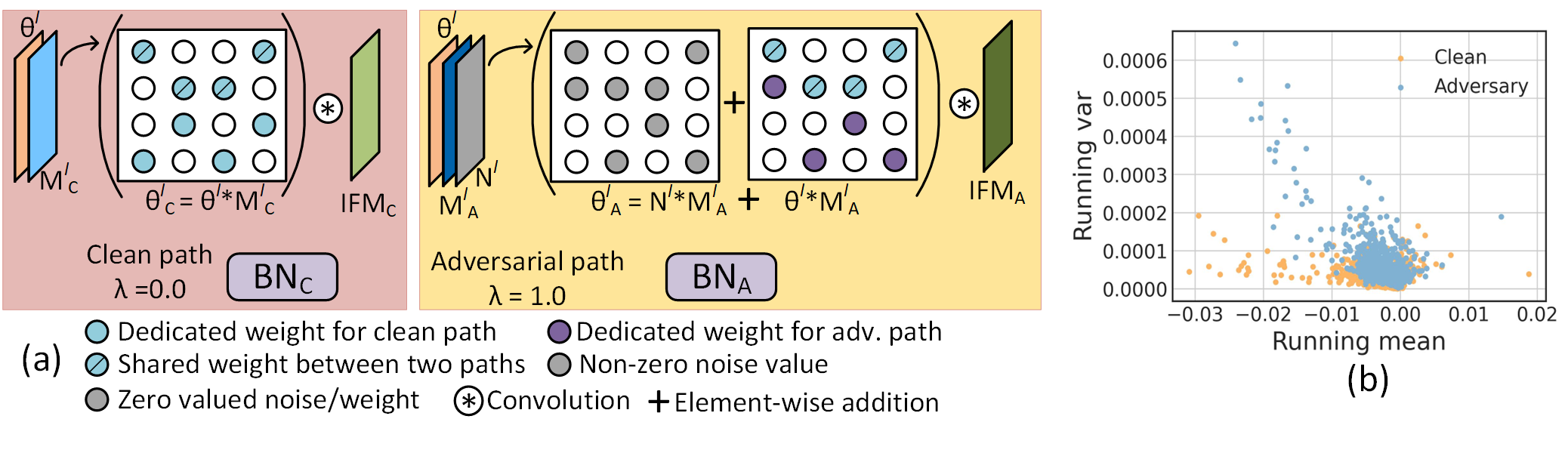}
\vspace{-6mm}
\centering
   \caption{(a) Dedicated convolutional weights in SMART. $M^l_A$ and $M^l_A$ are masks associated with clean and adversarial path selections that then convolve with the associated feature map, $IFM_C$ and $IFM_A$, respectively. For the initial layers $\mathcal{C}(M^l_C) = \mathcal{C}(M^l_A) = 1.0$, for later layers $\mathcal{C}(M^l_C) \ne \mathcal{C}(M^l_A) < 1.0$. Note, we follow \cite{kundu2022fast} to have dedicated BNs for each path. (b) BN statistics of two BN paths. Here, we used $2^{nd}$ BN of the $4^{th}$ basic block of ResNet34.}
\label{fig:chat_conditional_path}
\vspace{-4mm}
\end{figure}
\begin{table}[!t]
\caption{Comparison of different sparsity choices at the module level and across clean and adversarial expert path levels. We use ResNet34 (on CIFAR-10) which has four basic-block modules. We use `D' and `S' based on whether a specific module is kept dense or sparse, respectively. For sparse module CONVs, we equally partition the parameters setting density to 0.5, i.e. $\mathcal{C}(M^l_C) = \mathcal{C}(M^l_A) = 0.5$. }
\begin{center}
\scriptsize\addtolength{\tabcolsep}{-3.0pt}
\begin{tabular}{c|c|c|cc|cc}
\hline
Dataset & Basic-block & $\mathcal{C}(M^l_i)$ & \multicolumn{2}{c|}{Focus: Clean} & \multicolumn{2}{c}{Focus: Adversary} \\
{} & type & & \multicolumn{2}{c|}{($\lambda=0.0$)} & \multicolumn{2}{c}{($\lambda=1.0$)} \\
\cline{4-7}
{} &  {} &  & CA & RA  & CA & RA \\
\hline
\hline
{}       & [S S S S] & 0.0  & {95.20} & {1.18} & {43.20} & {9.68}\\
CIFAR-10       & [D D S S] & 0.0 & {95.22} & {16.18} & {84.60} & {51.56}\\
{}  & [D D S S] & 0.25 & \textbf{95.26} & \textbf{19.26} & \textbf{89.04} & \textbf{57.84} \\       
\hline
\end{tabular}
\end{center}
\label{tab:obs1_obs2}
\vspace{-4mm}
\end{table}
Inspired by the observation made in Section \ref{sec:motiv}, we now present sparse mixture once-for-all adversarial training (SMART), that require both fewer operations and parameters compared to the SOTA alternatives.

Formally, consider a $L$-layer DNN parameterized by $\bm{\Theta}$ and let $\bm{\theta}^l \in \mathbb{R}^{k^l \times k^l \times C^l_i \times C^l_o}$ represent the weight tensor of layer $l$. Here, $C^l_o$ and $C^l_i$ are the $\#$filters and $\#$channels per filter, respectively, with $k^l$ representing kernel height/width.
For each element $\theta^l$ in $\bm{\theta}^l$, for clean and adversarial images, we use its original and noisy variants, respectively. Moreover, drawing inspiration from findings of Section \ref{sec:motiv}, for each weight tensor, we introduce two sparsity masks, one associated with each boundary scenario of clean and adversary. Thus, the $l^{th}$-layer weight tensor for clean path is given by
\begin{align}
    \bm{\theta}^l_C = \theta^l \odot \bm{M}^l_c.
\end{align}
Here, $M^l_C$ represents the binary mask associated to clean path weights. For the adversarial path, we use a different binary mask $M^l_A$, and transform the weight via the addition of a noise tensor $\bm{\eta}^l \in \mathbb{R}^{k^l \times k^l \times C^l_i \times C^l_o}$ scaled by  $\alpha^l$, as follows
 \vspace{-2mm}
\begin{align}
    \bm{\theta}^l_A = (\bm{\theta}^l + \alpha^l\cdot \bm{\eta}^l)\odot \bm{M}^l_A; \; \; \; \eta^l \mathord{\sim} \mathcal{N}(0, (\sigma^l)^2).
    \label{eq:noise_trans}
\end{align}
$\bm{\eta}^l$ is formed from the normal distribution mentioned above, and we only add one scalar as a trainable parameter for each layer as $\alpha^l$. Note, $\sigma^l$ is the std deviation of weights at layer $l$. 

At later layers, we set the masks such that the parameter density for both the paths remain 0.5, meaning $\mathcal{C}(M^l_C) = \mathcal{C}(M^l_A) = 0.5$. $\mathcal{C}(.)$ is the density evaluation function measuring the fraction of non-zero elements in a tensor. Also, we define $M^l_i$ as $M^l_C \cap M^l_A$ highlighting the shared locations where both the masks have non-zero values. $\mathcal{C}(M^l_i) = 0$ means the two sparse paths to use mutually exclusive set of non-zero weights. At the earlier layers, due to high parameter utilization of the models we set $\mathcal{C}(M^l_C) = \mathcal{C}(M^l_A) = \mathcal{C}(M^l_i) = 1.0$, allowing all the parameters to be in both the expert paths. 

Inspired by \cite{he2019parametric}, for all the layers, we use the conditional weight transformation via noise (Eq. \ref{eq:noise_trans}) that is conditioned by a hyperparameter $\lambda$ to be either 0 or 1 to choose either of the two boundary conditions of clean and adversary. We select PGD-7 attack to generate perturbed image batches. The proposed SMART algorithm is detailed in Algorithm \ref{alg:smart}.

\textbf{Observation 1.} \textit{Sparse weight tensor at all layers throughout the model where for any layer $l$, $\mathcal{C}(M^l_C) = \mathcal{C}(M^l_A) = 0.5$, significantly degrade model performance for both the expert paths.}

As shown in Table \ref{tab:obs1_obs2}, row 1, the model performance is significantly inferior for both the boundary clean and adversary conditions, further strengthening the observation made in Section \ref{sec:motiv}. In other words, higher parameter utilization of earlier layers demands a high fraction of non-zero parameters. Thus, with dense initial layer basic-blocks, model performance has a significant boost, shown in row 2 of Table \ref{tab:obs1_obs2}.  

\textbf{Observation 2.} \textit{For a layer $l$, shared non-zero weights of two sparse masks, $\mathcal{C}(M^l_i) \ne 0.0$, can boost CA and RA.}

As we observe in row 2 and 3 of Table \ref{tab:obs1_obs2}, the model with $\mathcal{C}(M^l_i)= 0.25$, performs better than that with $\mathcal{C}(M^l_i) = 0.0$. Interestingly, the improvement is more prominent in yielding SOTA accuracy for both CA and RA, at $\lambda=1.0$ (focus:adversary). This may hint at the better generalization ability of the model when at least a part of it learns on both clean and adversary. Based on this observation, for any layer $l$, we design our random sparse mask such that $\mathcal{C}(M^l_i) \ne 0.0$, and keep the masks frozen throughout the training.
\begin{algorithm}[!t]
\footnotesize
\SetAlgoLined
\DontPrintSemicolon
\KwData{Training set $\bm{X} \mathord{\sim} D$, model parameters $\bm{\Theta}$, clean and adversarial sparse masks $\bm{M_C}$, $\bm{M_A}$, trainable noise scaling factor $\bm{\alpha}$, binary conditioning param. $\lambda$, mini-batch size $\mathcal{B}$.}
\textbf{Output:} trained model parameters $\bm{\Theta}$, $\bm{\alpha}$.\\
\For{$\text{i} \leftarrow 0$ \KwTo \KwTo {$ep$}}
{
    \For{$\text{j} \leftarrow 0$ \KwTo {${n_{\mathcal{B}}}$}}
    {
        $\text{Sample clean image-batch of size } \mathcal{B}/2$ ($\bm{X}_{0:{\mathcal{B}/2}}$, $\bm{Y}_{0:{\mathcal{B}/2}}$) $\mathord{\sim} D$\;
        $\mathcal{L}_C \leftarrow \texttt{computeLoss}({\bm{X}_{0:{\mathcal{B}/2}}}, {\bm{\Theta}}, {\bm{M_C}}, {\lambda}=0; \bm{Y}_{0:{\mathcal{B}/2}}) \text{ // condition to use weights w/o noise}$\;
        $\hat{\bm{X}}_{{\mathcal{B}/2}:{\mathcal{B}}} \leftarrow \texttt{createAdv}(\bm{X}_{{\mathcal{B}/2}:{\mathcal{B}}}, \bm{Y}_{{\mathcal{B}/2}:{\mathcal{B}}}) \text{ // adversarial image}$\;
        $\text{// creation}$\;
        $\mathcal{L}_A \leftarrow \texttt{computeLoss}({\hat{\bm{X}}_{{{\mathcal{B}/2}}: {\mathcal{B}}}}, {\bm{\Theta}}, {\bm{M_A}}, {\lambda}=1, \bm{\alpha}; \bm{Y}_{{{\mathcal{B}/2}}: {\mathcal{B}}}) \text{ // condition to use transformed weights}$\;
        $\mathcal{L} \leftarrow 0.5*\mathcal{L}_C + 0.5*\mathcal{L}_A$\;
        $\texttt{updateParam}(\bm{\Theta}, \bm{\alpha},\nabla_{\mathcal{L}})$\;
    }
}
 \caption{SMART Algorithm}
 \label{alg:smart}
\end{algorithm}
\begin{figure*}[!h]
\includegraphics[width=0.85\textwidth]{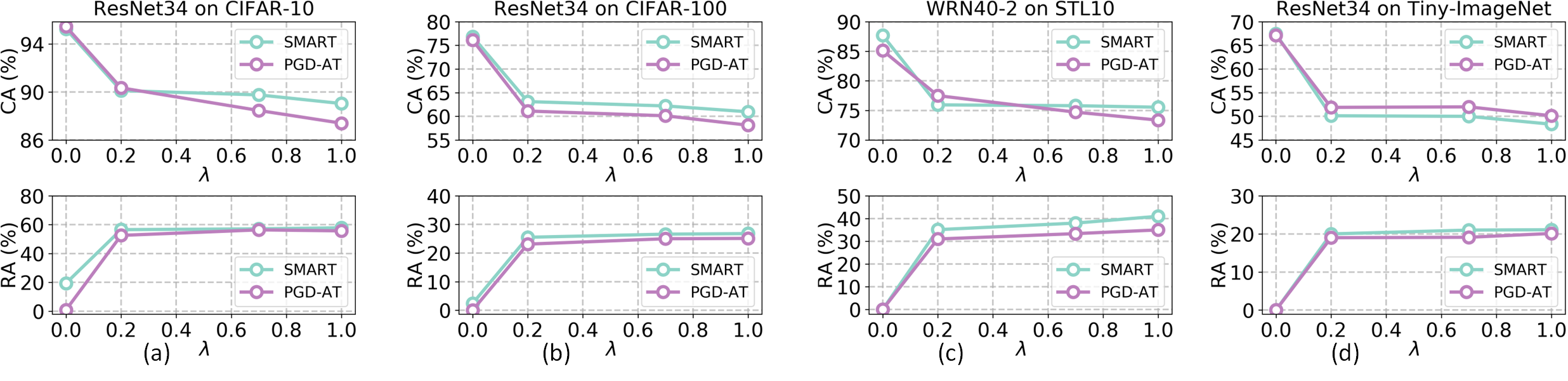}
\vspace{-6mm}
\centering
   \caption{Performance of models generated by SMART on (a) CIFAR-10, (b) CIFAR-100, (c) STL10, and (d) Tiny-ImageNet, for inference time $\lambda$ choices = [0.0, 0.2, 0.7, 1.0]. Importantly, for SMART, we get different trade-offs with one model trained once. However, for PGD-AT, we train a separate model for each $\lambda$-choice by changing its value in the loss described in Eq. \ref{eq:adv_loss}.}
\label{fig:results}
\vspace{-4mm}
\end{figure*}
\section{Experiments}
\label{sec:expt}

\subsection{Dataset, Model, and Training Setup}
We conducted experiments on CIFAR-10, CIFAR-100 \cite{krizhevsky2009learning}, Tiny-ImageNet \cite{hansen2015tiny} with ResNet-34 \cite{he2016deep}, and STL10 \cite{coates2011analysis} with WRN40-2 \cite{zagoruyko2016wide}. For CIFAR and STL10 we followed similar hyper-parameter settings as \cite{wang2020once}. For Tiny-ImageNet we used an initial learning rate of 0.1 with cosine decay and trained the model for 80 epochs. 
We used the PGD-$k$ attack for adversarial image generation during training and set  $\epsilon$ and $k$ to 8/255 and 7, respectively. As in \cite{he2019parametric}, we initialize the trainable noise scaling-factor $\alpha^l$ for a layer $l$ to $0.25$.

To measure the performance, we evaluated the trained models' CA and RA. CA is the classification accuracy on the original clean test images and RA is the accuracy on adversarially perturbed images generated from the test set. 

\subsection{Analysis of Results}
To perform inference on SMART models, we chose four  $\lambda$s as our test set. Note, during training we trained with only two $\lambda$ choices of [0,1]. During test, we leverage the noise re-scaling \cite{kundu2022fast} to tune the noise severity that gets added to the weights. In particular, we reduce the noise proportional to the value of $\lambda$. Also, similar to \cite{wang2020once, kundu2022fast}, we chose $BN_A$ for any test-time $\lambda > 0$. 
Fig. \ref{fig:results} shows SMART model to perform similar to that of PGD-AT trained models. These comparable performances at different CA-RA trade-off scenarios clearly showcase SMART as a once-for-all adversarial training method.  

\begin{table}[!t]
\caption{Comparison of SMART with exiting alternatives on CIFAR-10. Note, here we normalize the parameter cost with that of a baseline model with all the parameters present. For PGD-AT, the parameter cost (N) depends on the support $\lambda$ size, with a minimum value of 2. $\uparrow$, $\downarrow$, indicates the higher the better and the lower the better, respectively.}
\begin{center}
\scriptsize\addtolength{\tabcolsep}{-3.0pt}
\begin{tabular}{c|c|c|c|c|>{\columncolor{Gainsboro!60}}c}
\hline
Importance & Evaluation & PGD-AT  & OAT  & FLOAT  & SMART  \\
selection & metric & \cite{madry2017towards} & \cite{wang2020once} &  \cite{kundu2022fast} & (ours) \\
\hline
\hline
 Clean          & CA$\uparrow$ & 94.81 & 92.82 & 95.22            & \textbf{95.26} \\
($\lambda=0.0$) & RA$\uparrow$ & 0.84  & 21.72 & \textbf{21.97}   & 19.36 \\
                & Normalized params. $\downarrow$ & N$\times$ & 1.47$\times$ & 1$\times$ & 0.54$\times$\\
                & MAC overhead $\downarrow$ & 0 & 15M & 0 & 0 \\
                & Add overhead $\downarrow$ & 0 & 0 & 0 & 0 \\                
\hline
Adversary        & CA$\uparrow$ & 87.43  & 88.15 & 89.01 & \textbf{89.04} \\
($\lambda=1.0$) & RA$\uparrow$ & 55.72  & 53.67 & 56.31   & \textbf{57.84} \\
                & Normalized params. $\downarrow$ & N$\times$ & 1.47$\times$ & 1$\times$ & 0.54$\times$\\
                & MAC overhead $\downarrow$ & 0 & 15M & 0 & 0 \\
                & Add overhead $\downarrow$ & 0 & 0 & 21.28M & 11.4M \\
\hline
\end{tabular}
\end{center}
\label{tab:comaprison}
\vspace{-4mm}
\end{table}

\textbf{Comparison with other SOTA methods.} Table \ref{tab:comaprison} shows a detailed comparison of SMART with existing alternatives of PGD-AT, OAT, and FLOAT across various evaluation metrics. As we can see, SMART consistently provides similar or better accuracy/robustness while requiring only $0.54\times$ non-zero parameters compared to a standard ResNet34. In terms of extra additions, SMART requires $1.85\times$ fewer ops of only 11.84M ADD operations compared to FLOAT.

\subsection{Generalization on Other Attacks}
\begin{figure}[!h]
\includegraphics[width=0.40\textwidth]{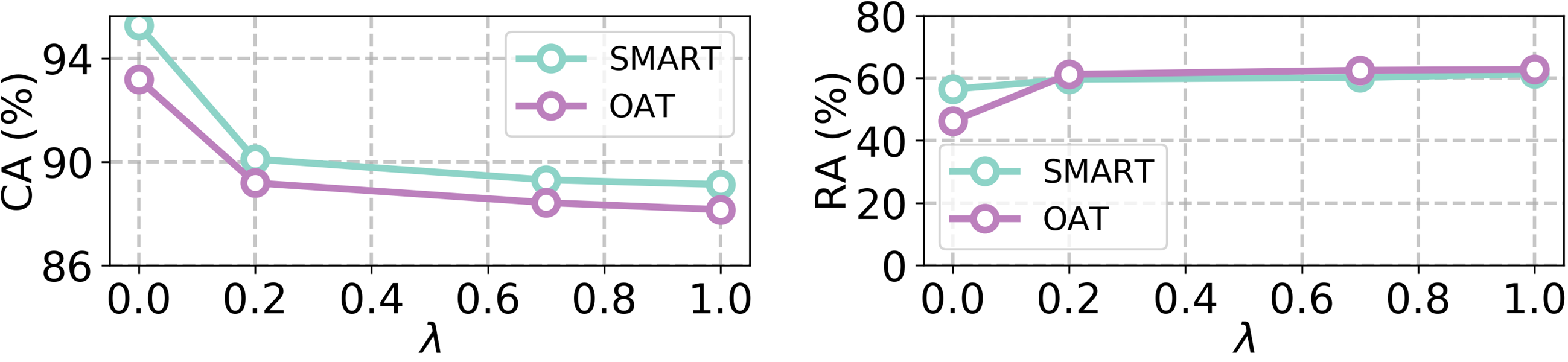}
\vspace{-4mm}
\centering
   \caption{Results of SMART on FGSM attacks at different CA-RA trade-off choices with $\lambda$, with ResNet34 on CIFAR-10.}
\label{fig:smart_fgsm}
\vspace{-6mm}
\end{figure}

To evaluate the generalization of SMART models, we used fast gradient sign method (FGSM) attack to generate image perturbations. As shown in Fig. \ref{fig:smart_fgsm}, SMART model generalizes similar to that of OAT, costing much fewer parameters or ops.
\section{Conclusions}
\label{sec:conc}
This paper presents a compute and parameter efficient weight conditioned learning of DNN models that can in-situ trade-off between accuracy and robustness. In particular, to effectively present reduced parameter conditioning, we first analyzed the layer-wise model parameter utility and then present SMART, a sparsity driven conditional learning that, at the later layers, learns two different sparse weight tensors for clean and adversarial images, respectively. More specifically, the later layers use two different set of sparse masks for clean and perturbed images. This not only, helps reduce the compute power for each path, but also helps each path learn better on its dedicated task. Extensive experiments showed SMART to outperform the existing SOTA, FLOAT by up to $1.53\%$ (on RA at $\lambda=1.0$) while saving additional compute by up to $1.85\times$.

\bibliographystyle{IEEEbib}
\small
\bibliography{biblio}

\end{document}